\documentclass{article}

\usepackage{arxiv}

\usepackage[utf8]{inputenc} % allow utf-8 input
\usepackage[T1]{fontenc}    % use 8-bit T1 fonts
\usepackage{hyperref}       % hyperlinks
\usepackage{url}            % simple URL typesetting
\usepackage{booktabs}       % professional-quality tables
\usepackage{amsfonts}       % blackboard math symbols
\usepackage{nicefrac}       % compact symbols for 1/2, etc.
\usepackage{microtype}      % microtypography
\usepackage{lipsum}

\usepackage{graphicx}
\usepackage{mathtools}
\usepackage{amsfonts}
\usepackage{calrsfs}
\DeclareMathAlphabet{\pazocal}{OMS}{zplm}{m}{n}

\usepackage{booktabs}
\usepackage{breqn}
\usepackage{xcolor}
\usepackage{hyperref}

\usepackage{float}

\usepackage{rotating}
\usepackage{multirow}
\usepackage{multicol}

\usepackage[utf8]{inputenc}

\usepackage{graphicx}
\usepackage{amsmath}

\usepackage[version=4]{mhchem}
\usepackage{siunitx}
\usepackage{longtable,tabularx}
\usepackage{bm}
\usepackage[inline]{enumitem}
\usepackage{array}

\title{PaDGAN: A Generative Adversarial Network for Performance Augmented Diverse Designs}
    
\author{
  Wei Chen \\
  Siemens Corporate Technology\\
  Princeton, NJ 08540 \\
  \texttt{chen.wei@siemens.com} \\
   \And
 Faez Ahmed \\
  Northwestern University\\
  Evanston, IL 10601 \\
  \texttt{faez@northwestern.edu} \\
}

\newcommand{\eg}{{\em e.g.}}
\newcommand{\etal}{{\em et~al.}}
\newcommand{\ie}{{\em i.e.}}
\newcommand{\etc}{{\em etc.}}
\newcommand{\RNum}[1]{\uppercase\expandafter{\romannumeral #1\relax}}

\usepackage[linesnumbered,ruled,vlined]{algorithm2e} 

\begin{document}

\maketitle    

%%%%%%%%%%%%%%%%%%%%%%%%%%%%%%%%%%%%%%%%%%%%%%%%%%%%%%%%%%%%%%%%%%%%%%
\begin{abstract}
{\it 
Deep generative models are proven to be a useful tool for automatic design synthesis and design space exploration. When applied in engineering design, existing generative models face three challenges: 1)~generated designs lack diversity and do not cover all areas of the design space, 2)~it is difficult to explicitly improve the overall performance or quality of generated designs, and 3)~existing models generally do not generate novel designs, outside the domain of the training data. In this paper, we simultaneously address these challenges by proposing a new Determinantal Point Processes based loss function for probabilistic modeling of diversity and quality. With this new loss function, we develop a variant of the Generative Adversarial Network, named ``Performance Augmented Diverse Generative Adversarial Network'' or PaDGAN, which can generate novel high-quality designs with good coverage of the design space. Using three synthetic examples and one real-world airfoil design example, we demonstrate that PaDGAN can generate diverse and high-quality designs. In comparison to a vanilla Generative Adversarial Network, on average, it generates samples with a $28\%$ higher mean quality score with larger diversity and without the mode collapse issue. Unlike typical generative models that usually generate new designs by interpolating within the boundary of training data, we show that PaDGAN expands the design space boundary outside the training data towards high-quality regions. The proposed method is broadly applicable to many tasks including design space exploration, design optimization, and creative solution recommendation.
}
\end{abstract}

% \section
% *{KEYWORDS}
% Diversity, Generative Design, Determinantal Point Process, Generative Adversarial Network, High Performance Designs, Design Exploration

%%%%%%%%%%%%%%%%%%%%%%%%%%%%%%%%%%%%%%%%%%%%%%%%%%%%%%%%%%%%%%%%%%%%%%
% \begin{nomenclature}
% \entry{SVS}{Design variety metric proposed in Shah }
% \end{nomenclature}

\setlength{\belowdisplayskip}{5pt} \setlength{\belowdisplayshortskip}{5pt}
\setlength{\abovedisplayskip}{5pt} \setlength{\abovedisplayshortskip}{5pt}

%%%%%%%%%%%%%%%%%%%%%%%%%%%%%%%%%%%%%%%%%%%%%%%%%%%%%%%%%%%%%%%%%%%%%%
\section{Introduction}

A designer wants good design solutions which are creative and meet the performance requirements. By manually and iteratively exploring design ideas using experience and design heuristics, the designers take the risks of 1)~wasting time on unfavorable or even invalid design candidates and 2)~not exploring as deeply as they might want to. 
% Generative design is a design exploration process which may reduce these risks. 
An ideal design space exploration tool should ensure that, with low cost, one can dive deep in the design space and explore all feasible design alternatives. 
% In an ideal world, designers would input design goals into a generative design software, along with parameters such as performance requirements, materials, manufacturing methods, and cost constraints. The software would explore all the possible permutations of a solution and quickly generate design alternatives. It would test and learn from past experience which designs work and which don't. 

% Deep generative models are a topic of enormous recent interest, providing a powerful class of tools for the unsupervised learning of probability distributions over manifolds.

While recent advances in machine learning assisted automatic design synthesis and design space exploration are promising, the current methods are still far from this ideal picture. To model a design space, researchers have used deep generative models like variational autoencoders (VAEs)~\cite{kingma2013auto} and generative adversarial networks (GANs)~\cite{goodfellow2014generative}, as they can learn the distribution of existing designs. The hope is that by learning an underlying \textit{latent space}, which can represent most designs, one can automatically synthesize many new designs from the low-dimensional \textit{latent vectors} and design exploration becomes more efficient due to the reduced dimensionality~\cite{chen2017design,chen2019synthesizing,chen2019aerodynamic}. 
However, unlike image generation tasks where these generative models are commonly applied, engineering design problems have one or more performance (or quality) measures.
The quality measures how well a design achieves its intended goals and is defined based on the specific problem. For example, beam design problems often define quality based on the compliance value (single-objective)~\cite{bendsoe_topology_2004} or both compliance and natural-frequency (multi-objective)~\cite{ahmed2016structural}. For aerodynamic design, quality can be defined as the lift-to-drag ratio~\cite{chen2019aerodynamic} or the inverse of the drag coefficient~\cite{shu20203d}.
% For aircraft designs, researchers define quality using factors like target height achieved by the airplane~\cite{zhang20193d} or inverse of the drag coefficient~\cite{shu20203d}.
Current state-of-the-art generative models have no mechanism of explicitly promoting high-quality design generation.
One may spend huge effort to train a generative model, only to find many generated designs are infeasible or do not meet design requirements. One way of working around this problem is to exclude low-quality data while training~\cite{shu20203d}. However, such an approach may affect model performance due to reduced training sample size. This creates a need to explicitly embed the quality measurement into a generative model, so that it can learn to generate high-quality designs by making use of full data and their quality measurements.
% These techniques are inspired by huge advances in computer vision applications, where GANs have successfully generated human faces of people who never existed before~\cite{karras2019style}. 

% On the other hand, as a generative design tool, topology optimization is widely used to search for an optimal solution given a specific objective function and boundary conditions~\cite{wang2003level,allaire2004structural}. However, it suffers from many local optima and is sensitive to initialization. For complex problems, due to the large search space and high dimensionality, it is possible that the topology optimization does not converge or the solution is non-manufacturable or does not meet desired physical properties. In that case, we need to explore other alternative solutions. Thus, there is a need for generating diverse design candidates with good coverage of the design space.
% However, current state-of-the-art deep learning methods or topology optimization tools are not designed to provide design alternatives which are both diverse (\ie, designs significantly different from each other) and have high-performance. 

In this work, we focus on addressing the problem of simultaneously maximizing diversity and quality of generated designs. Specifically, we develop a new loss function, based on Determinantal Point Processes (DPPs)~\cite{kulesza2012determinantal}, for generative models to encourage both high-quality and diverse design synthesis. Using this loss function, we develop a new variant of GAN, named PaDGAN. We show that it can generate high-quality new samples with a good coverage of the design space. More importantly, we found that PaDGAN can expand the existing boundary of the design space towards high-quality regions, which indicates its ability of generating novel high-quality designs.

With the ability of generating high-quality and diverse designs from a (reduced) latent representation, the proposed PaDGAN can then be used for improving the efficiency in design space exploration. While it is interesting to see how exploring the low-dimensional latent space of the PaDGAN can accelerate exploration or improve the performance of the optimal solution, we leave that to future work. In this paper, we focus on the architecture of PaDGAN and its performance in design synthesis.

\section{Background and Related Work}

Our work produces generative models that synthesize diverse designs from latent representations. There are primarily two streams of related research: 1)~design synthesis and 2)~diversity measurement. Within these two fields, we provide a brief background on two techniques we use in this paper~\textemdash~GANs and DPPs~\textemdash~and their applications in design. Readers interested in a more comprehensive understanding of their background are advised to read Kulesza \etal{}'s work~\cite{kulesza2012determinantal} for DPPs and the chapter on ``Deep Generative Models'' in Ref.~\cite{goodfellow2016deep}.

%%%%%%%%%%%%%%%%%%%%%%%%%%%%%%%%%%%%%%%%%%%%%%%%%%%%%%%%%%%%%%%%%%%%%%
\subsection{Deep Generative Model-Based Design Synthesis}

To achieve automatic design synthesis, past researchers have used approaches based on shape grammar~\cite{gmeiner2013spatial, konigseder2016improving, shea2005towards}, graph enumeration~\cite{herber2017enumeration,kamesh2017topological}, functional models~\cite{bryant2005concept}, analogy~\cite{chakrabarti2011computer}, and constraint programming~\cite{wyatt2012supporting,wijkniet2018modified}. These methods often need to encode expert knowledge as either grammar rules, functional basis, or constraints. In recent years, data-driven design synthesis has become increasingly popular. Different from traditional design synthesis methods, data-driven methods do not necessarily require expert knowledge and can learn to generate plausible new designs from a database~\cite{chen2015high, chen2017design, d2017nonlinear, d2018deep, chen2019aerodynamic}.

% Design synthesis methods can be classified into two groups: rule-based and data-driven. Rule-based synthesis (e.g., grammars-based design synthesis~\cite{gmeiner2013spatial, konigseder2016improving} requires labeling of the reference points or surfaces and defining rule sets, so that new designs are synthesized according to this hard-coded prior knowledge.
% They have been used for automated design generation in multiple ways, like developing biologically inspired algorithms for shape customization~\cite{ulu2015dms2015} or proposing new shape grammar methods for motorcycle design~\cite{whiting2017automated}. 

% Data-driven design synthesis, which is increasingly becoming more popular~\cite{dering2017generative}, learns rules from a database and generates plausible new designs with similar structure and function to existing ones. Dimensionality reduction techniques which allow inverse transformations from the latent space back to the design space are a commonly used data-driven design synthesis method. They have been used to synthesize new designs from latent variables in design applications ranging from airfoil synthesis to 3D models~\cite{chen2015high, chen2017design, d2017nonlinear, d2018deep, chen2019aerodynamic}. Broadly, they are used for applications including data augmentation, design space exploration, and design optimization. 

% \faez{Note to self: Discuss this with Wei if there exists more high level task categories in which ED folks have used GANs. }

In the last few years, deep generative models have gained traction, due to their ability to learn complex feature representations. The family of deep generative models contains various methods like the Boltzmann machines, deep belief networks, and differentiable generator networks like VAEs and GANs. VAEs and GANs are the two most commonly used deep generative models for solving engineering design problems. 
For example, they have been used in applications like design exploration~\cite{burnap2016estimating, chen2017design, chen2019synthesizing}, surrogate modeling~\cite{cunningham2019investigation}, and material microstructure design~\cite{cang2017scalable, yang2018microstructural}.

\paragraph{Applications of Deep Generative Models in Design Synthesis.}

Many design applications have huge collections of unstructured design data (CAD models, images, microstructures, \etc{}) with hundreds of features and multiple functionalities. To learn from these complex datasets, deep generative models have increasingly been employed. 
For instance, 
% In contrast to their focus on aesthetics, our goal is to generate designs which look different from each other and are also of high-quality. 
Chen~\etal{}~\cite{chen2019aerodynamic,chen2018bezier} proposed a B\'ezierGAN model for airfoil parameterization and synthesis, and demonstrated significantly faster convergence to the optimum when optimizing over the latent space. 
Yang~\etal{}~\cite{yang2018microstructural} used a GAN to generate microstructures and performed design optimization over the latent space.
Chen~\etal{}~\cite{chen2019synthesizing} proposed a hierarchical GAN architecture to synthesize designs with inter-part dependencies.
Oh~\etal{}~\cite{oh2019deep} integrated topology optimization and generative models to generate designs which are optimized for engineering performance. 
These methods either do not explicitly consider the quality of generated designs or use a separate optimization process to search for high-quality designs.
Burnap~\etal{}~\cite{burnap2019design} used a VAE to generate new highly-rated automotive images, which are aesthetically pleasing. Shu~\etal{}~\cite{shu20203d} proposed a GAN-based model to generate high-quality 3D designs, where they improve the quality of generated samples by retraining the model on an updated dataset with low performing designs removed. In contrast, our method improves the quality of generated designs while training the deep generative model, without retraining or discarding any samples in the training data. Also, to the best of our knowledge, there is no generative model that simultaneously encourages diversity and quality. While the methods we develop in this work are applicable to most deep generative models, we use GANs to demonstrate our results and will describe them next.

\paragraph{Generative Adversarial Networks.}

GANs~\cite{goodfellow2014generative} model a game between a generative model (\textit{generator}) and a discriminative model (\textit{discriminator}). The generative model maps an arbitrary noise distribution to the data distribution (\ie, the distribution of designs in our scenario), thus can generate new data; while the discriminative model tries to perform classification, \ie, to distinguish between real and generated data. The generator $G$ and the discriminator $D$ are usually built with deep neural networks. As $D$ improves its classification ability, $G$ also improves its ability to generate data that fools $D$. 
Thus, a vanilla GAN (standard GAN with no bells and whistles) has the following objective function, which comprises of a discriminator loss term and a generator loss term:
\begin{equation}
\min_G\max_D V(D,G) = \mathbb{E}_{\mathbf{x}\sim P_{data}}[\log D(\mathbf{x})] +  \mathbb{E}_{\mathbf{z}\sim P_{\mathbf{z}}}[\log(1-D(G(\mathbf{z})))],
\label{eq:gan_loss}
\end{equation}
where $\mathbf{x}$ is sampled from the data distribution $P_{data}$, $\mathbf{z}$ is sampled from the noise distribution $P_{\mathbf{z}}$, and $G(\mathbf{z})$ is the generator distribution. 
A trained generator thus can map from a predefined noise distribution to the distribution of designs. The noise input $\mathbf{z}$ is considered as the latent representation of the data, which can be used for design synthesis and exploration.

\paragraph{Problems in Using GANs for Design Synthesis.}

Learning in GANs can be difficult in practice, which may be one of the reasons that they are less widely used in design compared to VAEs. Despite an enormous amount of recent work in the machine learning community, GANs are notoriously unstable to train, and it has been observed that they often suffer from \textit{mode collapse}~\cite{salimans2016improved}, in which the generator network learns how to generate samples from a few modes of the data distribution but misses many other modes. For instance, when training on multiple categories of designs, a GAN model would sometimes generate designs only for a single category~\cite{mao2017least}. Recent approaches~\cite{bang2018mggan, srivastava2017veegan, chen2016infogan} tackled mode collapse in one of two different ways: 1)~modifying the learning of the system to reach a better convergence point; or 2)~explicitly enforcing the models to capture diverse modes or map back to the true-data distribution. Solutions to the mode collapse problem range from designing a reconstructor network in VEEGAN~\cite{srivastava2017veegan} to matching the similarity matrix of generated samples with data~\cite{elfeki2019gdpp}. In contrast, PaDGAN addresses the mode collapse problem implicitly by virtue of promoting generation of diverse solutions, which encourages samples to cover different modes, hence alleviating mode collapse.

% Identified problems we address: 
% 1) Mode collapse or 
% 2) lack of diversity. 
% 3) Explicitly model the performance of generated designs.

% While conceptually simple and fairly straightforward to compute, GANs suffer from a couple of subtle numerical and optimization issues when used in practical applications. 
% We review and solve these in Sec.~\ref{sec:methodology}.

%\subsection{Applications of generative design for Engineering Design problems}

%%%%%%%%%%%%%%%%%%%%%%%%%%%%%%%%%%%%%%%%%%%%%%%%%%%%%%%%%%%%%%%%%%%%%%
\subsection{Measuring Design Coverage}

Massive highly redundant sources of audio, video, speech, text documents, and sensor data have become commonplace and are expected to become larger and more preponderant in the future~\cite{dube2016customer}. This brings a need to measure diversity of a set of items, such that redundancy in data can be reduced and machine learning models can be trained using data with a smaller sample size and which are not biased in favor of a few classes.
% DPPs belong to the class of Strongly Rayleigh (SR) measures; these measures benefit from the strongest characterization of negative association between similar items; as such, SR measures have benefited from significant interest in the mathematics community and more recently in machine learning. This, combined with their tractability, makes DPPs a particularly attractive tool for subset selection in machine learning, and is one of the key motivations for our work.
\textit{Diversity} (also called coverage or variety) is a measure of how different a set of items are from each other. Quantitatively, it is measured using two predominant ways~\textemdash~submodular functions or DPPs. Submodular functions are set functions with diminishing marginal gain property, which naturally model notions of coverage and diversity. They achieved among the top results on common automatic document summarization benchmarks (\eg, at the Document Understanding Conference~\cite{Lin2011class}).
In design, too, researchers have used submodular functions-based diversity measures to understand design space exploration using terms like \textit{variety}~\cite{shah2000evaluation, fuge2013automatically, ahmed2019measuring}. These functions have helped designers sift through large sets of ideas by ranking them ~\cite{Ahmed2017ranking} or selecting a diverse subset~\cite{ahmed2016discovering}. Ahmed~\etal{}~\cite{Ahmed2017ranking} compared DPPs~\cite{kulesza2012determinantal} with certain commonly used submodular functions. They concluded that unlike submodular functions, DPPs are more flexible, since they only need a valid similarity kernel as an input rather than an underlying Euclidean space or clusters. In this paper, we will use DPPs as a measure of diversity, which we will describe next.

\paragraph{Determinantal Point Processes.}

DPPs, which arise in quantum physics, are probabilistic models that model the likelihood of selecting a subset of diverse items as the determinant of a kernel matrix. 
Viewed as joint distributions over the binary variables corresponding to item selection, DPPs essentially capture negative correlations and provide a way to elegantly model the trade-off between often competing notions of quality and diversity. 
%Determinantal Point Processes (DPPs) are probabilistic models over subsets of a ground set that elegantly model the trade-off between often competing notions of quality and diversity. 
The intuition behind DPPs is that the determinant of a kernel matrix roughly corresponds to the volume spanned by the vectors representing the items. Points that ``cover'' the space well should capture a larger volume of the overall space, and thus have a higher probability. 
% DPPs also have a geometric intuition, where increasing the norm of a vector (quality of an item in the set) or increasing the angle between the vectors (dissimilarity between items in the set) increases the volume spanned by the volume spanned by the vectors~\cite{kulesza2012determinantal}.
As shown by Kulesza \etal \cite{kulesza2011k}, one of DPPs' advantages is that computing marginals, computing certain conditional probabilities, and sampling can all be done in polynomial time. In this paper, we focus on another advantage of DPPs, which is the decomposition of DPP kernels into quality and similarity terms.

For the purposes of modeling real data, the most relevant construction of DPPs is through L-ensembles~\cite{borodin2009determinantal}.
An L-ensemble defines a DPP via a positive semi-definite matrix $L$ indexed by the elements of a subset $S$.
The kernel matrix $L$ defines a global measure of similarity between pairs of items, so that more similar items are less likely to co-occur.
The probability of a set $S$ occurring under a DPP is calculated as:
\begin{equation}
\mathbb{P}_L(S) = \frac{\det(L_S)}{\det(L+I)},
\label{eq:dppk}
\end{equation}
where $L_S \equiv [L_{ij}]_{ij \in S}$ denotes the restriction of $L$ to the entries indexed by elements of $S$, $I$ is an $N \times N$ identity matrix, and $N$ is the total number of items.
For any set size, the most probable subset under a DPP will have the maximum likelihood over $\mathbb{P}_L(S)$ or (equivalently) the highest determinant (the denominator can be ignored for maximizing determinant of a fixed set size). 
% DPPs also have a geometric intuition. Let $q_i$ , $q_j$ be two feature vectors of $\phi$ such that the DPP kernel verifies L = $\phi \phi ^T$; then $\sqrt{P(i,j)} \propto \text{Vol}(\phi_i \phi_j)$. 
%DPPs also have a geometric intuition, where increasing the norm of a vector (quality of an item in the set) or increasing the angle between the vectors (dissimilarity between items in the set) increases the volume spanned by the volume spanned by the vectors~\cite{kulesza2012determinantal}.
% As the similarity between two items increases, the probabilities of sets containing both of them decrease.
% Unlike many machine learning tools, DPPs are more flexible, since we only need to provide a valid similarity kernel (\eg, image or shape kernels), rather than explicit vectors representing each design in an underlying vector space.
Similar to sub-modular functions, one of the main applications of DPP is extractive document summarization, where it provided state-of-the-art results. In Section~\ref{sec:methodology}, we show how the decomposition of DPP kernels can be used to design a DPP-based loss function, which promotes quality and diversity of generated samples in a generative model.
%They have recently been used in design for subset selection~\cite{ahmed2016discovering} and ranking of ideas~\cite{Ahmed2017ranking}.

% \subsection{What are some of the recent methods to improve GANs?}

%%%%%%%%%%%%%%%%%%%%%%%%%%%%%%%%%%%%%%%%%%%%%%%%%%%%%%%%%%%%%%%%%%%%%%
\subsection{Comparison with State-of-the-Art and Our Contributions}

The work closest to ours is the GDPP method~\cite{elfeki2019gdpp} by Elfeki~\etal{}.
The authors devised an objective term that encourages the GAN to synthesize data with diversity similar to the training data. 
%They introduced a new penalty term to match the eigenvalues and eigenvectors of the generated fake data DPP kernel with their corresponding real data DPP kernel. 
PaDGAN differs from their method in three aspects. 
First, PaDGAN is stable against scaling of data while on validating GDPP for multiple test problems, we found that their method does not work for problems with training data at different scales. 
%Specifically, suppose we train the model on dataset A. Now if we multiply all points in dataset A by a constant (say ten) and train the GDPP again, it fails to generate samples similar to the training data. 
%2) our goal is to maximize quality and diversity while GDPP mimics the quality and diversity of the training data; 
Second, while PaDGAN aims to maximize the diversity of generated samples, GDPP aims to achieve a similar diversity value as the training data. By avoiding the goal of mimicking the diversity of the training data, PaDGAN will generate diverse samples even when the original training dataset is biased in favor of a few modes, while GDPP is designed to mimic the bias in generated samples. %As we show in Section 5, this helps generate more diverse new samples than the training data.
%we define quality using a performance estimator (\eg, a simulator or a surrogate model), while GDPP uses closeness to the training data as a quality measure.
Finally, whereas we maximize the quality of generated samples, whereas GDPP does not have such consideration. This feature of PaDGAN is helpful for design exploration as it can help discover novel high-quality designs (demonstrated in Section~\ref{sec:airfoil}).
%Additionally, we use a surrogate model to estimate quality, while GDPP uses distance from nearest point in the training data as a measure of quality.

% On further investigation, we found the reason behind this strange behavior arises from their loss function, which comprises of three parts: a vanilla GAN loss, diversity magnitude loss, and a diversity structure loss. The authors scale the third part to be between 0 to 1 (which measures similarity between eigen vectors), while the second part (difference in eigen values of generated and true samples) scales with the scaling of the training data. Hence, when we change the scale of the data, then their GAN loss ignores the third part, and fails to generate diverse samples. In contrast, PaDGAN is agnostic to scaling of the data as we scale the similarity between 0 and 1.

The scientific contributions and novelty of this work are as follows:
\begin{enumerate}

    \item We propose a novel design synthesis method that simultaneously encourage synthesis of diverse and high-performance designs.
    %We provide the first generative adversarial network to learn from the data how to automatically synthesize high-performance designs, which are also dissimilar from other generated designs.
    %Simultaneous exploration of design and performance spaces when learning to synthesize designs.
    %\item Integration of performance augmented DPP loss with the existing GAN architecture.
    
    \item We find that PaDGAN can expand the design space boundary towards high-quality regions that it had not seen from existing data.
    %We find that PaDGAN can expand the design space to generate novel high-quality designs, which it had not seen before.
    
    \item We propose a way to control the trade-off between quality and diversity in DPPs. Our method extends past work on decomposing a DPP kernel by providing a way to tune the relative importance of quality over diversity.
    
    \item We provide easy-to-verify test cases and metrics to validate any generative models, whose goal is to maximize sample quality and/or coverage over a dataset with multiple modes.
    
    % \item While we demonstrated the efficacy of DPP-based loss function for GAN models, we claim that the method generalizes to other existing generative design architectures like VAEs, leading them to generate performance augmented diverse designs.
        
    %A new method to balance quality and diversity in Determinantal Point Processes 
    % \faez{@Wei, I think this is also a contribution?}
    % \wei{@Faez, I don't think we have any specific way of balancing quality and diversity.}
    % \faez{we raise quality to the power lambda, before adding quality terms to the diagonal of the kernel matrix.}
    % \wei{Not sure if this counts as a contribution since it is quite trivial.}
    
\end{enumerate}

%%%%%%%%%%%%%%%%%%%%%%%%%%%%%%%%%%%%%%%%%%%%%%%%%%%%%%%%%%%%%%%%%%%%%%

\begin{figure*}[!ht]
\centering
\includegraphics[width=1\textwidth]{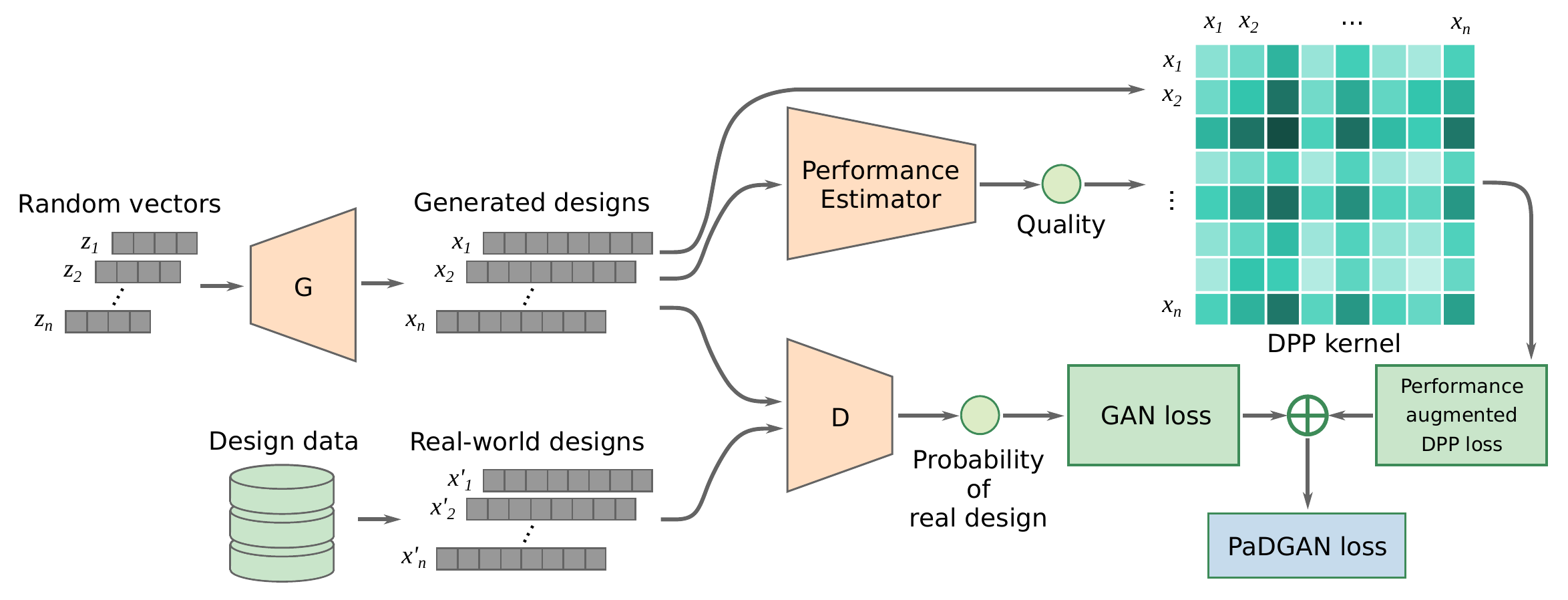}
\caption{Architecture of PaDGAN.}
\label{fig:architecture}
\end{figure*}

\begin{figure}[!ht]
\centering
\includegraphics[width=0.6\textwidth]{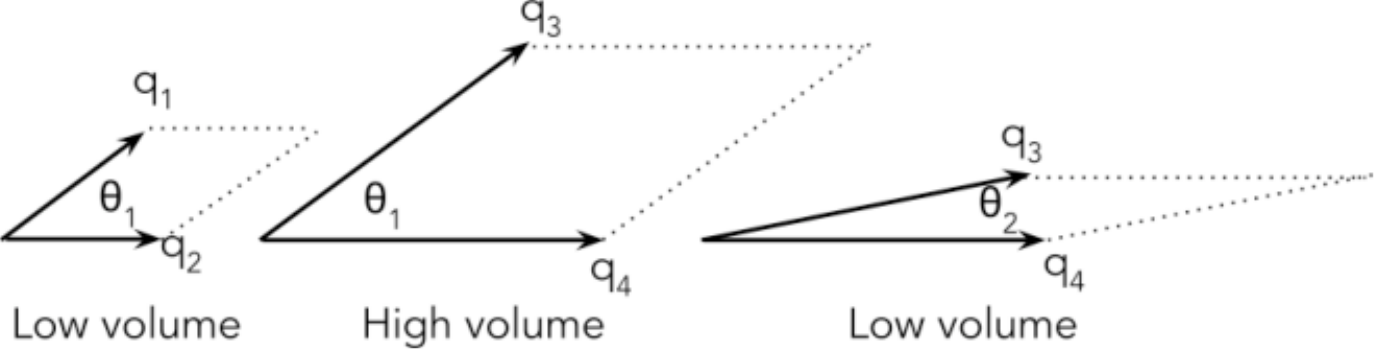}
\caption{Visualization of volume in 2-D space, where $q_i$ represent the quality of an item and $\theta_i$ shows how similar they are. Comparing the leftmost figure to the central figure, we observe that the similarity is same but the quality magnitude increases, leading to a higher volume. From the central figure to the rightmost figure, we observe that the quality magnitude is the same, but the similarity increases, leading to a lower volume (or diversity) encompassed by the two vectors. %This reduces the probability of the two items getting selected.
}
\label{fig:volume}
\end{figure}

\section{Methodology}
\label{sec:methodology}
Built on a standard GAN architecture, PaDGAN introduces a \textit{performance augmented DPP loss} which measures the diversity and quality of a batch of generated designs during training. The overall model architecture of PaDGAN is shown in Fig.~\ref{fig:architecture}. In this section, we begin by describing how to decompose a DPP kernel, then proceed on how to create a DPP loss which augments high performing designs, and finally provide a method to balance diversity and quality using a quality dial. We also add a note on improving training stability at the end. 
%We explain this new loss and the training method in the following sections.

%%%%%%%%%%%%%%%%%%%%%%%%%%%%%%%%%%%%%%%%%%%%%%%%%%%%%%%%%%%%%%%%%%%%%%
\subsection{Decomposition of a DPP kernel}

DPP kernels can be decomposed into quality and diversity parts~\cite{kulesza2012determinantal}. 
% A positive semi-definite DPP kernel $L$ can be written as a Gram matrix, and decomposed into $L = B^T~
% B$ (using methods like Cholesky decomposition), where a row of vector $B$ can be considered a vector representation of each item. 
% The $(i,j)^{th}$ entry of a positive semi-definite DPP kernel $L$
% the kernel matrix $L$ is the dot product of row $i$ and column $j$ of matrix $B$. 
The $(i,j)^{th}$ entry of a positive semi-definite DPP kernel $L$ can be expressed as:
\begin{equation}
 L_{ij} = q_i\;\phi(i)^T\;\phi(j)\;q_j.
 \label{eq:L_ij}
\end{equation}

We can think of $q_i \in R^+$ as a scalar value measuring the quality (or performance) of an item $i$, and $\phi(i)^T\;\phi(j)$ as a signed measure of similarity between items $i$ and $j$. The decomposition enforces $L$ to be positive semidefinite. 
% This decomposition enforces $L$ to be positive semidefinite and allows us to independently model quality and diversity, and then combine them into a unified model. 
Suppose we select a subset $S$ of samples, then this decomposition allows us to write the probability of this subset $S$ as the square of the volume spanned by $q_i \phi _i$ for $i \in S$ using the equation below:
\begin{equation}\label{eq:eq4}
    \mathbb{P}_L(S)~\propto~\prod_{i \in S} ({q_i}^2) \det(K_S),
\end{equation}
where $K_S$ is the similarity matrix of $S$.

The first term increases with the quality of the selected items, and the second term increases with the diversity of the selected items. 
%We will refer to $q$ as the quality model and $\phi$ as the diversity model. 
As item $i$’s quality $q_i$ increases, so do the probabilities of sets containing item $i$. As two items $i$ and $j$ become more similar, ${\phi_i}^T \phi_j$ increases and the probabilities of sets containing both $i$ and $j$ decrease. From a geometric intuition, the determinant of $L_Y$ is equal to the squared volume of the parallelepiped spanned by the vectors $q_i\phi_i$ for $i \in Y$. We show an illustration of this intuition in Fig.~\ref{fig:volume}.
The magnitude of the vector representing item $i$ is $q_i$, and its direction is $\phi_i$. It shows how DPPs decomposed into quality and diversity naturally balance the two objectives of high-quality and high diversity. 

When selecting a subset $S$ of items, without the diversity term, we would choose high-quality items, but we would tend to choose similar high-quality items over and over. Without the quality term, we would get a very diverse set, but we might fail to include the most important items in $S$, focusing instead on low-quality outliers. By combining the two models, we can achieve a more balanced result. The key intuition of PaDGAN is that if we can find a way to add the term from Eq.~(\ref{eq:eq4}) to the objective function of any generative model, then while training it will be encouraged to generate high probability subsets, which will be both diverse and high-quality. In the next section, we define such a loss function.

While, the authors used this decomposition to find quality and similarity terms from a known kernel, we reverse this procedure to create the kernel $L$ for a sample of points generated by PaDGAN from known inter-sample similarity values and quality. 
%The quality or performance values can be estimated by an external model. 
Note that in a DPP model, the quality or performance of an item is a scalar value, like compliance, displacement, drag-coefficient, \etc{} The quality can be estimated using an external model (like a physics-based simulator) or by finding the distance of current performance of a design from a target performance. For multi-dimensional cases, quality can be derived by taking the norm of multiple dimensions.
The similarity terms $\phi(i)^T \phi(j)$ can be derived using any similarity kernel, which we represent using $k(\mathbf{x}_i,\mathbf{x}_j) = \phi(i)^T \phi(j)$ and $\| \phi(i) \| = \| \phi(j) \| = 1$.
Here $\mathbf{x}_i$ is a vector representation of a design.

% Note that for vanilla DPPs there are very few restrictions on how the quality term is calculated. 
% One can run a complex physics based simulation model to calculate the quality, use surrogate models or derive the quality score by learning subjective human expert preferences. 
% However, we will impose two practical restrictions on the quality model in the later sections~\textemdash~it should be computable in a manageable time, which is needed to train a neural network model on large datasets and we should be able to calculate the gradient of the quality, to integrate it with the GAN objective function. 

%%%%%%%%%%%%%%%%%%%%%%%%%%%%%%%%%%%%%%%%%%%%%%%%%%%%%%%%%%%%%%%%%%%%%%
\subsection{Performance Augmented DPP Loss}

Our performance augmented DPP loss models diversity and quality simultaneously and gives a lower loss to sets of designs which are both high-quality and diverse. 
Specifically, we construct a kernel matrix $L_B$ for a generated batch $B$ based on Eq.~(\ref{eq:L_ij}). For each entry of $L_B$, we have
\begin{equation}
L_B(i,j) = k(\mathbf{x}_i,\mathbf{x}_j)\left(q(\mathbf{x}_i)q(\mathbf{x}_j)\right)^{\gamma_0},
\label{eq:L_B}
\end{equation}
where $\mathbf{x}_i,\mathbf{x}_j \in B$, $q(\mathbf{x})$ is the quality function at $\mathbf{x}$, and $k(\mathbf{x}_i,\mathbf{x}_j)$ is the similarity kernel between $\mathbf{x}_i$ and $\mathbf{x}_j$. We add $\gamma_0$ term as a dial to control the weight of quality, which is further explained in Section~\ref{sec:dial}.

The performance augmented DPP loss is expressed as
\begin{equation}
\pazocal{L}_{\text{PaD}}(G) = -\frac{1}{|B|}\log\det(L_B) = -\frac{1}{|B|}\sum_{i=1}^{|B|} \log\lambda_i,
\label{eq:pad_loss}
\end{equation}
where $\lambda_i$ is the $i$-th eigenvalue of $L_B$. By adding this loss to the vanilla GAN's objective from Eq.~(\ref{eq:gan_loss}), the problem becomes:
\begin{equation}
\min_G\max_D V(D,G) + \gamma_1 \pazocal{L}_{\text{PaD}}(G),
\label{eq:overall_loss}
\end{equation}
where $\gamma_1$ controls the weight of $\pazocal{L}_{\text{PaD}}$(G). To update any weight $\theta_G^i$ in the generator in terms of $\pazocal{L}_{\text{PaD}}(G)$, we descend its gradient based on the chain rule:
\begin{equation}
\frac{\partial \pazocal{L}_{\text{PaD}}(G)}{\partial \theta_G^i} = \sum_{j=1}^{|B|} \left( \frac{\partial \pazocal{L}_{\text{PaD}}(G)}{\partial q(\mathbf{x}_j)} \frac{d q(\mathbf{x}_j)}{d \mathbf{x}_j} + \frac{\partial \pazocal{L}_{\text{PaD}}(G)}{\partial \mathbf{x}_j} \right) \frac{\partial \mathbf{x}_j}{\partial \theta_G^i},
\label{eq:gradient}
\end{equation}
where $\mathbf{x}_j = G(\mathbf{z}_j)$.

Equation~(\ref{eq:gradient}) indicates a need for $dq(\mathbf{x})/d\mathbf{x}$, which is the gradient of the quality function. In practice, this gradient is accessible when the quality is evaluated through any performance estimator that is differentiable, like adjoint-based solver methods. If the gradient of a performance estimator is not available, one can either use numerical differentiation or approximate the quality function using a differentiable surrogate model (\eg, a neural network-based surrogate model). In our experiments in Section~\ref{sec:airfoil}, we use a neural network-based surrogate model. We will explore the possibility of using an automatic differentiation enabled simulator (\eg, an adjoint solver) as the performance estimator in future studies.

%%%%%%%%%%%%%%%%%%%%%%%%%%%%%%%%%%%%%%%%%%%%%%%%%%%%%%%%%%%%%%%%%%%%%%
\subsection{Introducing a quality dial for DPP kernels}
\label{sec:dial}

Note that we modified the original objective to introduce $\gamma_0$ as a parameter. We found that traditional DPP decomposition does not allow us to change the importance of quality versus diversity within a given kernel. This means that if we fix the quality scores and similarity scores, the trade-off between the two cannot be controlled.
A na\"ive way to increase the importance of quality would be to multiply the quality scores by a large constant and expect it to increase its importance relative to diversity. 
However, with careful observation one would realize that this approach would not work. Using the geometric interpretation of the DPPs, this would be equivalent to scaling all lengths by the same factor, which will not affect the volumes relative value. As quality and diversity objectives are multiplied together to get the probability of the set (Eq.~(\ref{eq:eq4})), to change the relative importance, we need to adjust the dynamic range of the quality scores. We do this by using an exponent to change the distribution of quality. When $\gamma_0 = 0$, all quality scores collapse to one and the resultant PaDGAN model only generates diverse designs. In contrast, for large values of $\gamma_0$, the highest quality scores have the largest probability mass and PaDGAN only generates the highest quality designs, ignoring diversity. This method of balancing diversity and quality provides more flexibility to PaDGAN and in general, can be used for many applications of DPPs.

%%%%%%%%%%%%%%%%%%%%%%%%%%%%%%%%%%%%%%%%%%%%%%%%%%%%%%%%%%%%%%%%%%%%%%
\subsection{Improving PaDGAN stability}
%Escalating $\gamma_1$ to improve training stability

Stabilization of GAN learning remains an open problem and in this section, we provide a heuristic method to improve GAN stability, when using a surrogate model for evaluating quality. 
Note that in Eq.~(\ref{eq:gradient}), the quality gradient is used in the back propagation step. If the quality gradients are not accurate, the generator learning can go astray.
This is not a problem when the quality estimator is a simulator that can reasonably evaluate (even with low-fidelity) any design in the design space, irrespective of the designs being invalid or unrealistic.
However, it creates problems when we use a surrogate model. A surrogate model is normally trained only on realistic designs and hence may perform unreliably on unrealistic ones. In the initial stages of training, a GAN model will not always generate realistic designs during training. This makes it difficult for the surrogate model to correctly guide the generator's update and may cause stability issues.
To avoid this problem, we propose two small modifications to PaDGAN:
\begin{enumerate}
\item \textit{Realisticity weighted quality}. Specifically, we weight the predicted quality at $\mathbf{x}$ by the probability of $\mathbf{x}$ being the real design (predicted by the discriminator):
$$q(\mathbf{x}) = D(\mathbf{x})q'(\mathbf{x})$$
where $q'(\mathbf{x})$ is the predicted quality (by a surrogate model for example), and $D(x)$ is the discriminator's output at $\mathbf{x}$.
\item An \textit{escalating schedule} for setting $\gamma_1$ (the weight of DPP loss). A GAN is more likely to generate unrealistic designs in its early stage of training. Thus, we initialize $\gamma_1$ at 0 and increase it during training, so that PaDGAN focuses on learning to generate realistic designs at the early stage, and takes quality into consideration later when the generator can produce more realistic designs. The schedule is set as:
$$\gamma_1 = \gamma_1'\left(\frac{t}{T}\right)^p$$
where $\gamma_1'$ is the value of $\gamma_1$ at the end of training, $t$ is the current training step, $T$ is the total number of training steps, and $p$ is a factor controlling the steepness of the escalation.
\end{enumerate}

We can also consider the uncertainty of the quality estimation and put a lower weight on the quality score when the uncertainty is high. However, we only consider the above two modifications in this paper and leave others to future work. Note that these modifications are only needed if one is using a performance estimator (\eg, a surrogate model) which gives unreliable quality predictions for unrealistic designs.

%%%%%%%%%%%%%%%%%%%%%%%%%%%%%%%%%%%%%%%%%%%%%%%%%%%%%%%%%%%%%%%%%%%%%%
\section{Experiment}

So far, we have shown how the mathematical components of PaDGAN will encourage it to generate high-quality and diverse samples. In this section, we will describe experiments, which can help us validate our claims. These experiments are carefully designed such that the outcome of any generative models can be verified easily. 
This section introduces the experimental settings for each example. To show the merit of modeling quality and diversity simultaneously, we compare the PaDGAN with alternative models where those two attributes are modeled separately. 
% We compare PaDGAN against three other models. The first is a vanilla GAN model, which is commonly used in most Machine Learning applications. The second is $GAN_D$ model, which uses the objective function of PaDGAN, but ignores the quality of items and maximizes only diversity. The third is $GAN_Q$ model, which rewards high-quality items but ignores diversity. 
In the following sections, we show that for three multi-modal synthetic problems, PaDGAN outperforms all other methods by achieving both high-quality and high diversity. Finally, after showing that the claims hold on three test cases, we apply PaDGAN on a real-world airfoil synthesis problem. We find that PaDGAN can discover new regions of high-quality designs, which are outside the design domain over which it was trained.

% We demonstrate the effectiveness of the PaDGAN via three synthetic examples and a real-world example. 

%%%%%%%%%%%%%%%%%%%%%%%%%%%%%%%%%%%%%%%%%%%%%%%%%%%%%%%%%%%%%%%%%%%%%%
\subsection{Data and Quality Measure}

% Mode collapse is quantified by the number of real modes recovered in fake data, and the generation quality is quantified by the \% of High-Quality Samples.

\begin{figure}[!ht]
\centering
\includegraphics[width=0.5\textwidth]{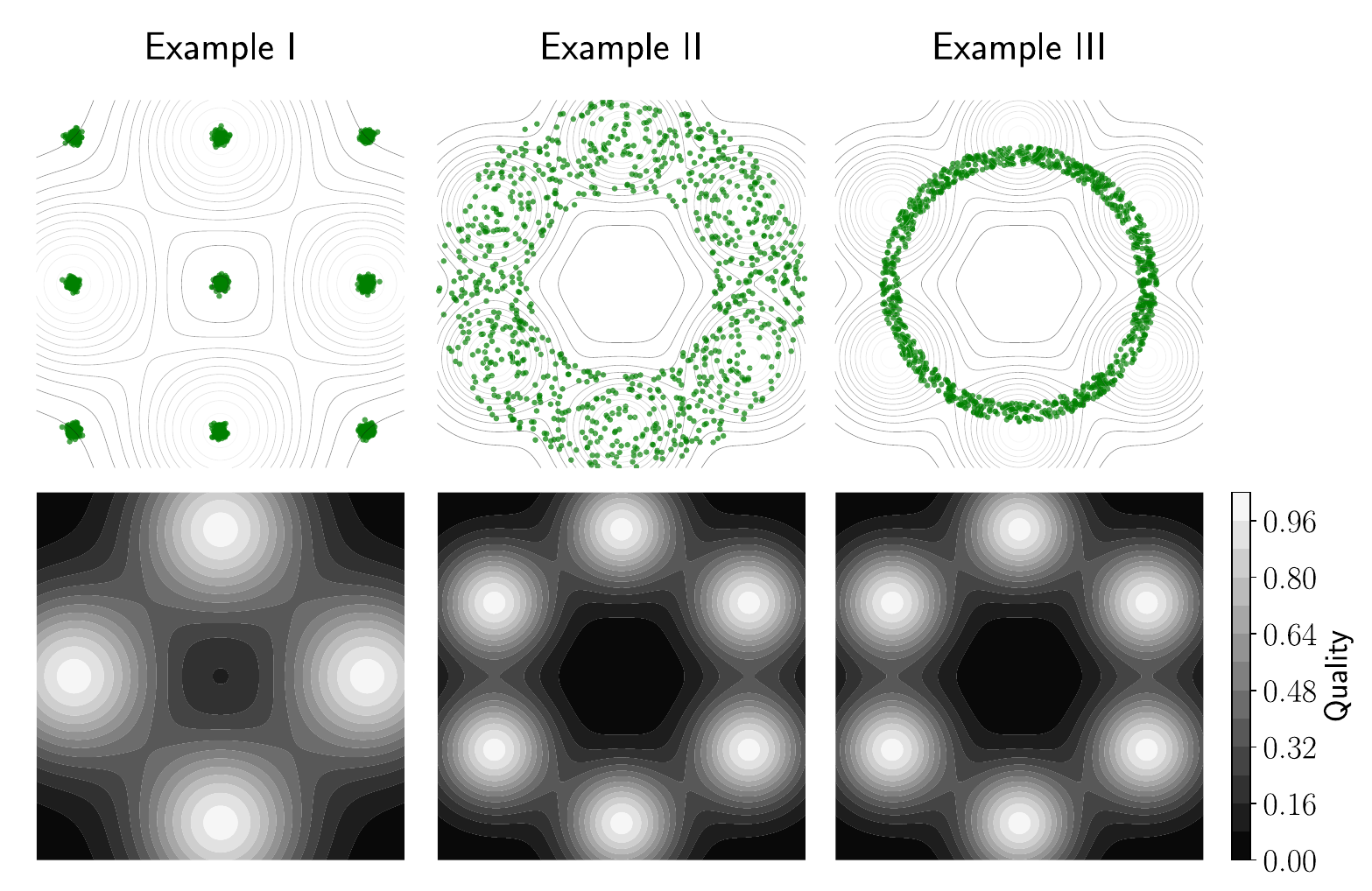}
\caption{Data and quality functions in synthetic examples. The green dots in the top plots represent data and the contours represent quality functions.}
\label{fig:data}
\end{figure}

\paragraph{Synthetic example \RNum{1}.} 
The purpose of creating 2D synthetic examples is to test the performance of PaDGAN given known ground truth and visualize the results in terms of diversity and quality. These examples are analogical to any 2D design problem, where designs are represented by two variables. In this synthetic example \RNum{1}, we generate a ring-shaped dataset, with data uniformly distributed between two origin-centered circles of 0.25 and 0.5 in radius, respectively (Fig.~\ref{fig:data}). We use a density function of an unnormalized Gaussian mixture as the quality function:
\begin{equation}
q(\mathbf{x}) = \sum_{k=1}^K \exp \left(-\frac{(\mathbf{x}-\mu_k)^T(\mathbf{x}-\mu_k)}{2\sigma^2}\right),
\label{eq:gaussian_mix}
\end{equation}
where $\mu_k$ is the mode of the $k$-th mixture component and $\sigma$ is the standard deviation. The centers $\mu_1,...,\mu_K$ are evenly spaced around a circle centered at the origin and with a radius of 0.4. We set $K=6$ and $\sigma\approx 0.1$. 
Hence, there are six peaks of quality and points are evenly spread between two concentric circles in the training data. 
Ideally, by simultaneously maximizing diversity and quality, we expect generating more samples near the six local optima (\ie, modes) of the quality function, and those samples should be spread out and evenly distributed among all six mixture components.

\paragraph{Synthetic example \RNum{2}.} The data in this example are nine clusters placed on a $3\times 3$ grid (Fig.~\ref{fig:data}). Similar to synthetic example \RNum{1}, we use Eq.~(\ref{eq:gaussian_mix}) as the quality function. Here we set $K=4$ and $\sigma\approx 0.16$. Four out of nine clusters (modes) of the data overlap with local optima of the quality function.
We expect that if both diversity and quality are considered, the generator should produce most samples in all the four high-quality clusters and few samples in other clusters (instead of generating most samples from a single high-quality cluster).

\paragraph{Synthetic example \RNum{3}.} This example is the same as example \RNum{1}, except that data is bounded within two origin-centered circles of 0.325 and 0.375 in radius, respectively (Fig.~\ref{fig:data}). The purpose of decreasing the coverage of data is to demonstrate PaDGAN's capability of extrapolating in the high-quality regions (\ie, expanding the boundary of existing design space towards high-quality regions). 

\paragraph{Airfoil example.} An airfoil is the cross-sectional shape of a wing or a propeller/rotor/turbine blade. In this example, we use the UIUC airfoil database\footnote{\url{http://m-selig.ae.illinois.edu/ads/coord_database.html}} as our data source. It provides the geometries of nearly 1,600 real-world airfoil designs. Each design is represented by discrete 2D coordinates along their upper and lower surfaces. We preprocessed and augmented the dataset based on Ref.~\cite{chen2019aerodynamic} to generate a dataset of 38,802 airfoils. The lift to drag ratio $C_L/C_D$ is a common objective in aerodynamic design optimization problems. Thus we used $C_L/C_D$ as the performance measure, which can be computed using XFOIL software~\cite{drela1989xfoil}. To provide the gradient of the quality function for Eq.~(\ref{eq:gradient}), we trained a neural network-based surrogate model on all 38,802 airfoils to approximate the quality. Note that for all the examples, we scaled the quality scores between 0 and 1.
We show a subset of $100$ randomly chosen example airfoils from the training data in the left plot of Fig.~\ref{fig:airfoils_tsne}. 
% Glassware
% If we have space and time, we can add this example too.  Here, we define a quality function, say menthe volume held in the glass. THe volume should be above a threshold. Next, we train the GAN to generate glasses which hold atleast that much volume and are of diverse shapes.

\begin{figure*}[!ht]
\centering
\includegraphics[width=1\textwidth]{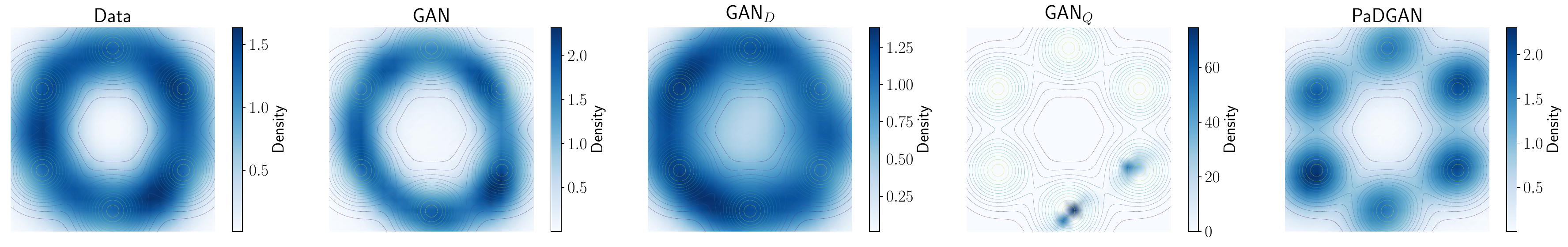}
\caption{Results on synthetic example \RNum{1}. The leftmost plot show the density of training data. The rest of the plots show the density of samples generated by different models.}
\label{fig:Donut2D_MixRing6_density}
\end{figure*}

\begin{figure*}[!ht]
\centering
\includegraphics[width=1\textwidth]{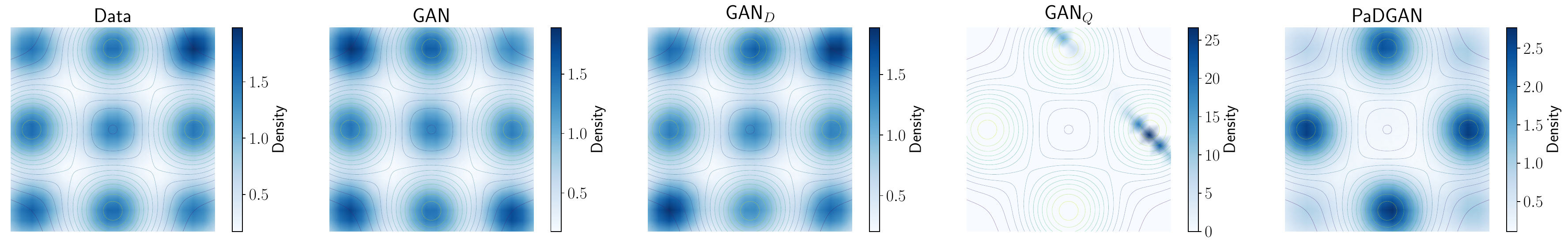}
\caption{Results on synthetic example \RNum{2}. The leftmost plot show the density of training data. The rest of the plots show the density of samples generated by different models.}
\label{fig:Grid2D_MixRing4_density}
\end{figure*}

\begin{figure*}[!ht]
\centering
\includegraphics[width=1\textwidth]{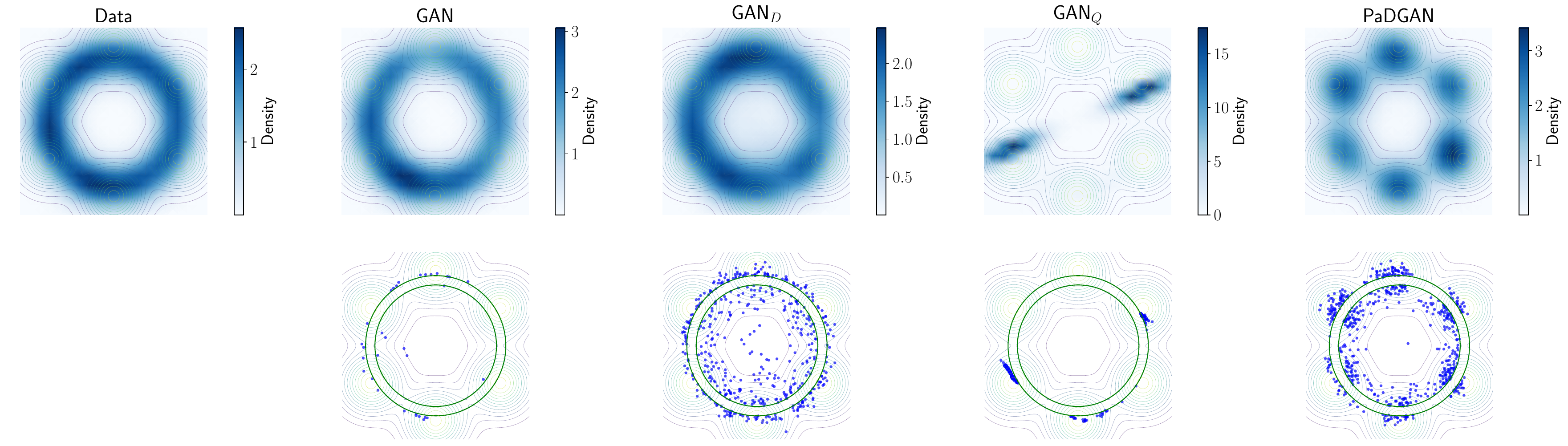}
\caption{Results on synthetic example \RNum{3}. The leftmost plot show the density of training data. The remaining plots in the top row show the density of samples generated by different models. We observe that GAN and $GAN_D$ generate samples similar to the data, while $GAN_Q$ suffers with mode collapse and only generates samples for two clusters. PaDGAN generates more samples in the high quality region, ignoring the low-quality areas of the training data. Plots in the bottom row show only the generated samples (blue dots) which are ``outside'' the region of training data (indicated by two green circles). We observe that PaDGAN generates unseen data in the high-quality areas while other methods either do not explore outside the domain or generate low-quality samples. 
}
\label{fig:ThinDonut2D_MixRing6_expand_density}
\end{figure*}

%%%%%%%%%%%%%%%%%%%%%%%%%%%%%%%%%%%%%%%%%%%%%%%%%%%%%%%%%%%%%%%%%%%%%%
\subsection{Model Configuration and Training}

To demonstrate the effectiveness of the PaDGAN, we compare it with the following three models:
\begin{enumerate}
    \item GAN: a vanilla GAN with the objective of Eq.~(\ref{eq:gan_loss}).
    \item GAN$_D$: PaDGAN with $\gamma_0=0$ in Eq.~(\ref{eq:L_B}), \ie, which only optimizes for diversity and ignores the quality. 
    \item GAN$_Q$: a vanilla GAN which ignores diversity and only optimizes for quality using the following additional term $\pazocal{L}_Q(G)=-\frac{1}{|B|}\sum_{i=1}^{|B|} q(\mathbf{x}_i)$. The training objective is then set to:
    $$\min_G\max_D V(D,G) + \gamma_2 \pazocal{L}_Q(G)$$
    where $\gamma_2$ controls the weight of the quality objective.
\end{enumerate}

To find similarity between designs, we use a RBF kernel with a bandwidth of 1.0 when constructing $L_B$ in Eq.~(\ref{eq:L_B}), \ie, $k(\mathbf{x}_i,\mathbf{x}_j)=\exp(-0.5\|\mathbf{x}_i-\mathbf{x}_j\|^2)$. This gives a value between 0 to 1, with a higher value for more similar designs. In synthetic examples, we set $\gamma_0=2$ and $\gamma_1=0.5$ for PaDGAN and $\gamma_2=10$ for GAN$_Q$ (these settings were chosen based on a few initial trials and domain knowledge). The generators and discriminators are fully connected neural networks. In the airfoil example, we set $\gamma_0=2$ and $\gamma_1=0.2$ for PaDGAN. We used a residual neural network (ResNet)~\cite{he2016deep} as the surrogate model and a B\'ezierGAN~\cite{chen2019aerodynamic,chen2018bezier} to generate airfoils. For simplicity, we refer to the B\'ezierGAN as a vanilla GAN and the B\'ezierGAN with loss $\pazocal{L}_{PaD}$ as a PaDGAN in the airfoil example in the rest of the paper. Detailed network architecture and hyperparameter settings can be found in our open-source code~\footnote{\url{https://github.com/wchen459/PaDGAN}}.

%%%%%%%%%%%%%%%%%%%%%%%%%%%%%%%%%%%%%%%%%%%%%%%%%%%%%%%%%%%%%%%%%%%%%%
\subsection{Evaluation}

We use the \textit{diversity score} and the \textit{quality score} of generated samples to measure the performance of generative models. The diversity score is expressed as the mean log determinant of the similarity matrix:
\begin{equation}
s_{\text{div}} = \frac{1}{n}\sum_{i=0}^n \log\det(L_{S_i}),
\label{eq:div_score}
\end{equation}
where $n$ is the number of times diversity is evaluated, $S_i\subseteq Y$ is a random subset of $Y$ (the set of generated samples), and $L_{S_i}$ is the similarity matrix of $S_i$ with entries $L_{S_i}(j,k)=k(\mathbf{x}_j,\mathbf{x}_k)$ for each $\mathbf{x}_j,\mathbf{x}_k \in S_i$. The quality score is computed by taking the average quality of generated samples:
\begin{equation}
s_{\text{qa}} = \frac{1}{|Y|}\sum_{i=0}^{|Y|} q(\mathbf{x}_i),
\label{eq:qa_score}
\end{equation}
where $\mathbf{x}_i\in Y$ is a randomly generated design.

For synthetic examples, we define the \textit{overall score}, to measure the overall performance by combining measures for diversity and quality of generated samples:
\begin{equation}
s_{\text{overall}} = -\sum_k \frac{m_k}{|Y|}\log\left(\frac{m_k}{|Y|}\right),
\label{eq:overall_score}
\end{equation}
where $m_k$ is the number of generated samples within the one-sigma interval of the $k$-th mixture component of the quality function. The overall score is affected by both the amount of high-quality samples and the spread of those samples. The highest score occurs when there are the same number of generated samples within the one-sigma interval of each mixture component and no samples are outside those intervals.

In the experiments, we set $|Y|=1000$, $|S_i|=10$, and $n=1000$. To take into consideration the stochasticity of the model training, for each type of model (PaDGAN, GAN, GAN$_D$, and GAN$_Q$), we train them ten times for each experimental setting, and report the performance statistics for all those ten models (Figs.~\ref{fig:Donut2D_MixRing6_scores}, \ref{fig:Grid2D_MixRing4_scores}, and \ref{fig:airfoil_scores}). We report and discuss the results in the next section.

%%%%%%%%%%%%%%%%%%%%%%%%%%%%%%%%%%%%%%%%%%%%%%%%%%%%%%%%%%%%%%%%%%%%%%
\section{Results and Discussion}

In this section, we compare the performance of PaDGAN to its alternatives (\ie, GAN, GAN$_D$, and GAN$_Q$) and discuss the implication of these results.

%%%%%%%%%%%%%%%%%%%%%%%%%%%%%%%%%%%%%%%%%%%%%%%%%%%%%%%%%%%%%%%%%%%%%%
\subsection{Synthetic Examples}

Figures~\ref{fig:Donut2D_MixRing6_density}, \ref{fig:Grid2D_MixRing4_density}, and \ref{fig:ThinDonut2D_MixRing6_expand_density} show the density plots of generated samples for each model, which represents their \textit{generative distribution}. Ideally, when we sample designs from the generator, we want these designs to have a good coverage over real-world designs (\ie, the training data) and most of them should have high-quality. In Fig.~\ref{fig:Donut2D_MixRing6_density}, the generative distribution learned by a vanilla GAN fails to cover the entire training data (non-uniform contours). However, in both examples, the generative distribution of GAN$_D$ has a good coverage of the training data due to its diversity objective. This shows that the diversity objective by itself is capable of avoiding mode collapse. By replacing the diversity objective with a quality objective, GAN$_Q$ only generates samples near one of the optima of the quality functions, ignoring the others. In practice, this will give many high-quality samples but they all look very similar to each other. In contrast, the generative distribution of PaDGAN exhibits has a higher density near high-quality regions and also good coverage of the design space.

Figure~\ref{fig:ThinDonut2D_MixRing6_expand_density} shows that both GAN$_D$ and PaDGAN expands the boundary of training data. Particularly, PaDGAN expands the boundary towards high-quality regions. If these samples represent designs, it basically indicates that PaDGAN can expand the boundary of existing designs. We will further demonstrate this with a real design problem later. This promising result shows that by diversifying generated samples, PaDGAN is capable of expanding the design space towards the direction of high-quality regions. Note that this is not only filling the ``holes'' of the design space by interpolation, but also \textit{extrapolation} on the right direction. It is not surprising that the generator knows which direction to expand since it receives from the performance estimator the information of quality gradients. 

\begin{figure}[!ht]
\centering
\includegraphics[width=0.7\textwidth]{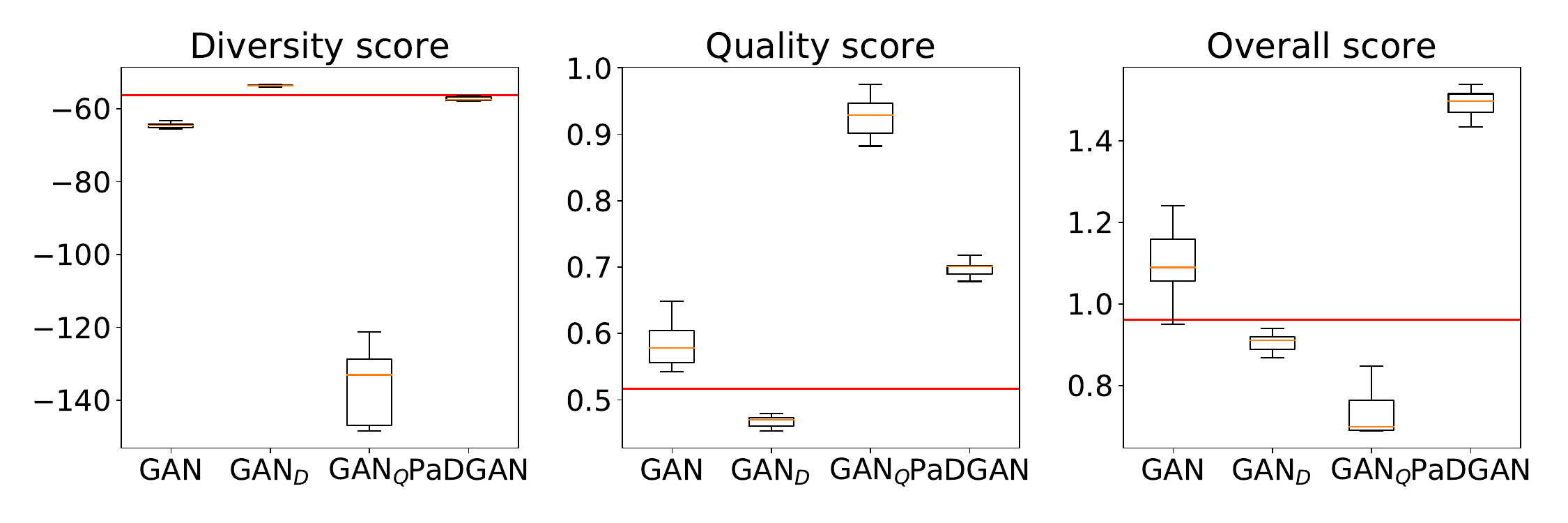}
\caption{Scores for synthetic example \RNum{1}. The red horizontal line denotes the diversity/quality score of the training data. The box plots show the statistics of ten models for each method.}
\label{fig:Donut2D_MixRing6_scores}
\end{figure}

\begin{figure}[!ht]
\centering
\includegraphics[width=0.7\textwidth]{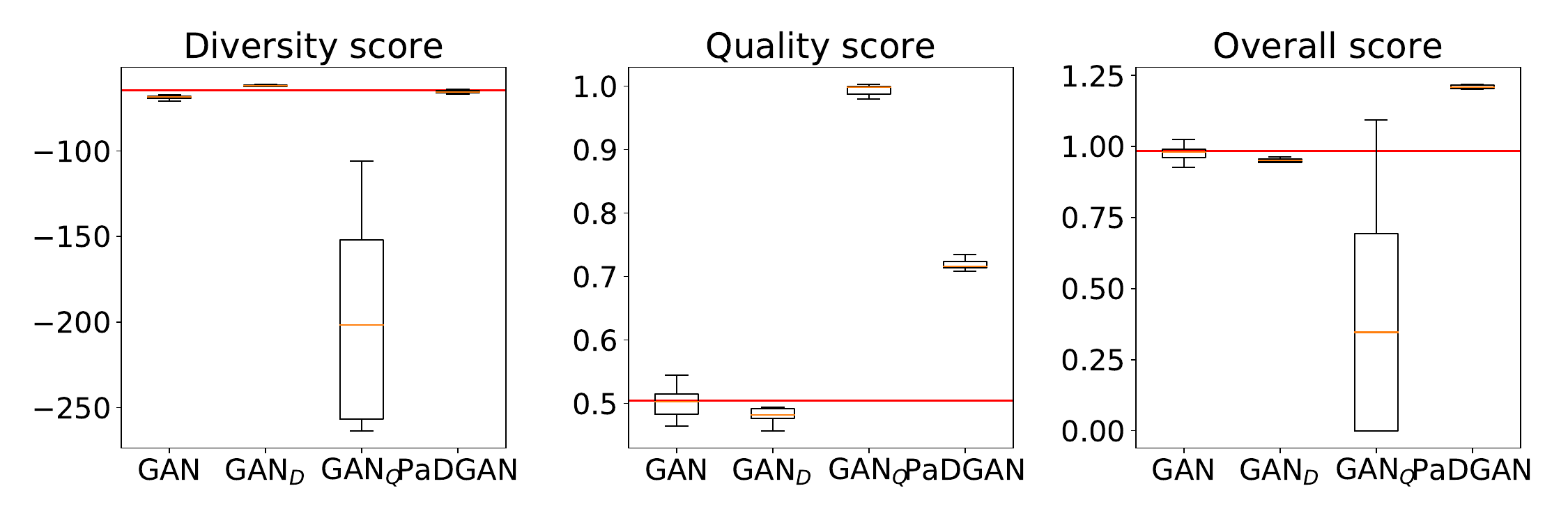}
\caption{Scores for synthetic example \RNum{2}. The red horizontal line denotes the diversity/quality score of the training data. The box plots show the statistics of ten models for each method.}
\label{fig:Grid2D_MixRing4_scores}
\end{figure}

\begin{figure}[!ht]
\centering
\includegraphics[width=0.7\textwidth]{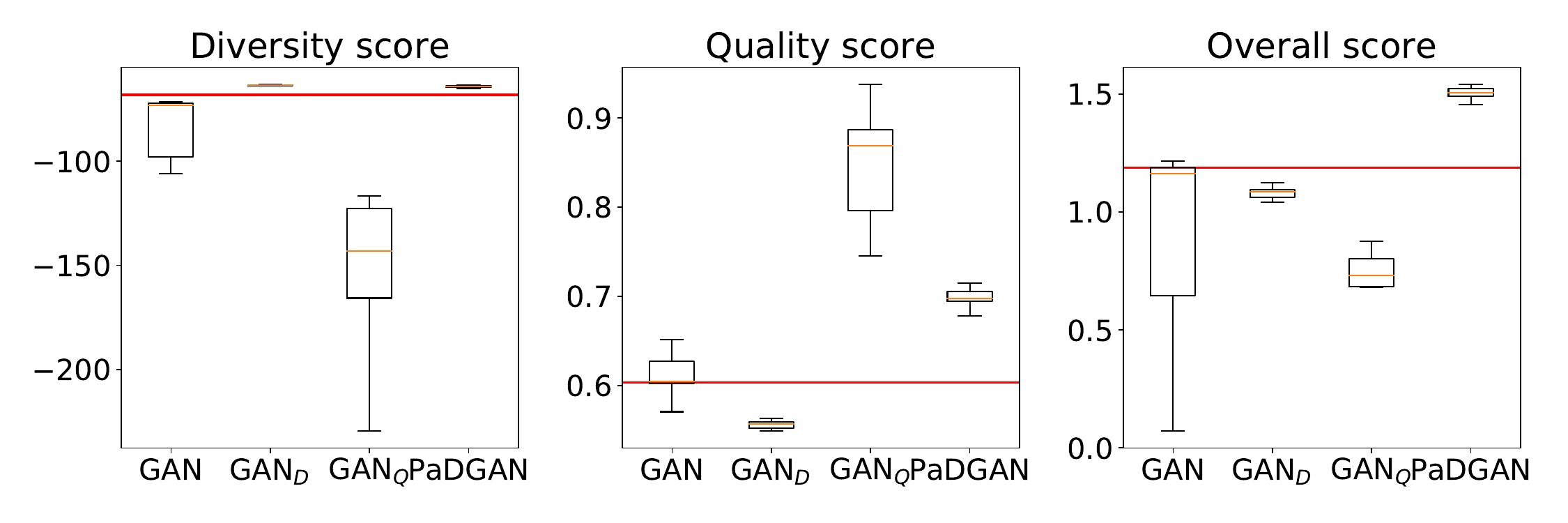}
\caption{Scores for synthetic example \RNum{3}. The red horizontal line denotes the diversity/quality score of the training data. The box plots show the statistics of ten models for each method.}
\label{fig:ThinDonut2D_MixRing6_scores}
\end{figure}

Figures~\ref{fig:Donut2D_MixRing6_scores}, \ref{fig:Grid2D_MixRing4_scores}, and \ref{fig:ThinDonut2D_MixRing6_scores} show the statistics of ten trained models for each method. Both figures tell that GAN$_D$ has the best performance in the diversity score and the worst performance in the quality score. GAN$_Q$ generates the highest quality samples, but has the lowest diversity scores, showing that all the samples very similar to each other. PaDGAN has the highest overall score in both examples, which shows that it generates high-quality samples that spread over different optima. The lowest variance indicates a consistent performance over multiple runs of PaDGAN training.

\subsection{Airfoil Example}\label{sec:airfoil}

\begin{figure*}[!ht]
\centering
\includegraphics[width=0.9\textwidth]{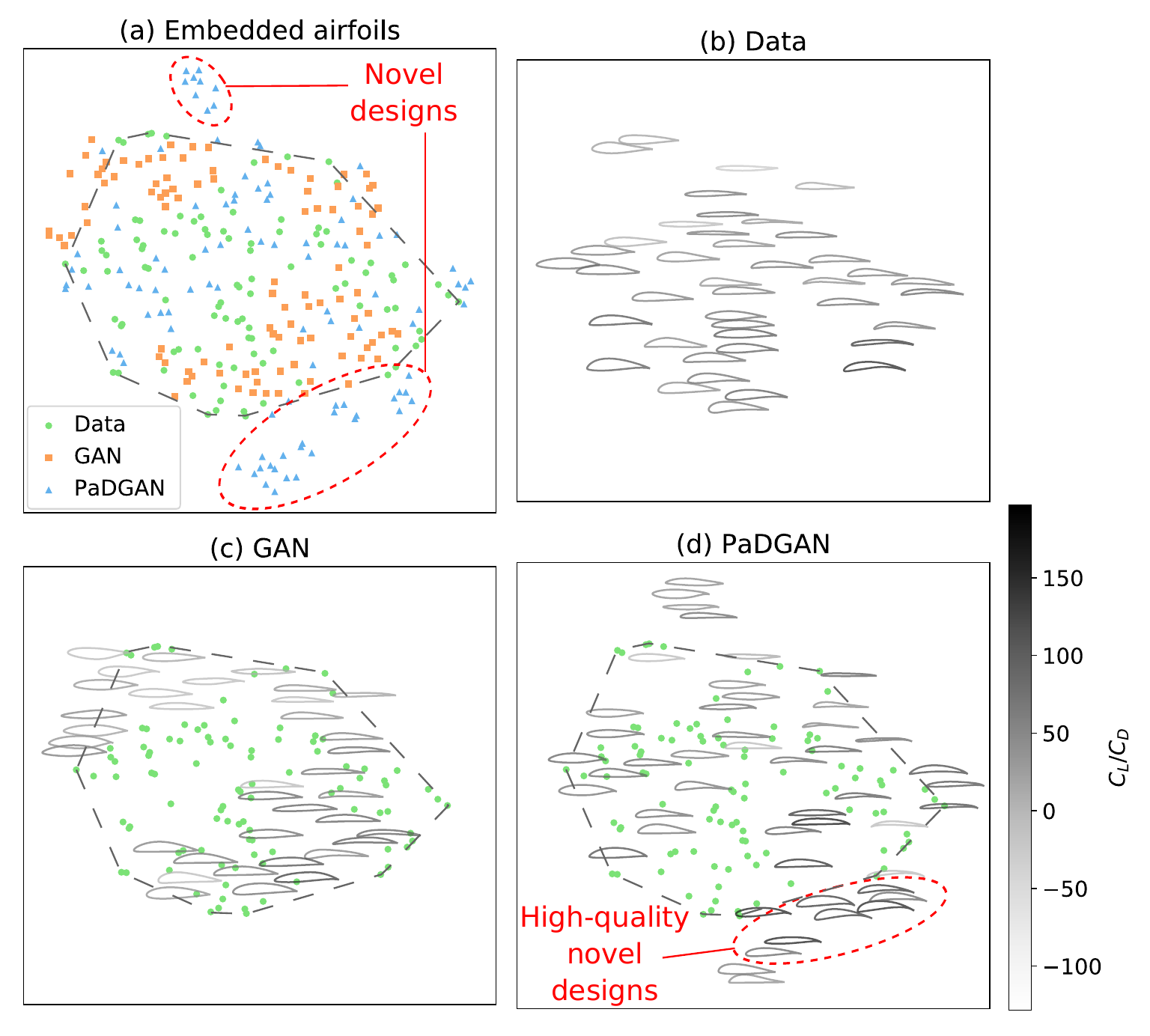}
\caption{To compare the distribution of real and synthetic airfoils, we map airfoil designs sampled randomly from training data, vanilla GAN, and PaDGAN through t-SNE into the same 2D space (shown in (a)). Plots (b)-(d) visualizes the airfoil geometries, where the shades represent quality (\ie, $C_L/C_D$). The dots in (c) and (d) represent training data. We label the convex hull of the sampled training data in Plots (a), (c), and (d), which roughly indicates the boundary of the original design space.}
\label{fig:airfoils_tsne}
\end{figure*}

\begin{figure}[!ht]
\centering
\includegraphics[width=0.35\textwidth]{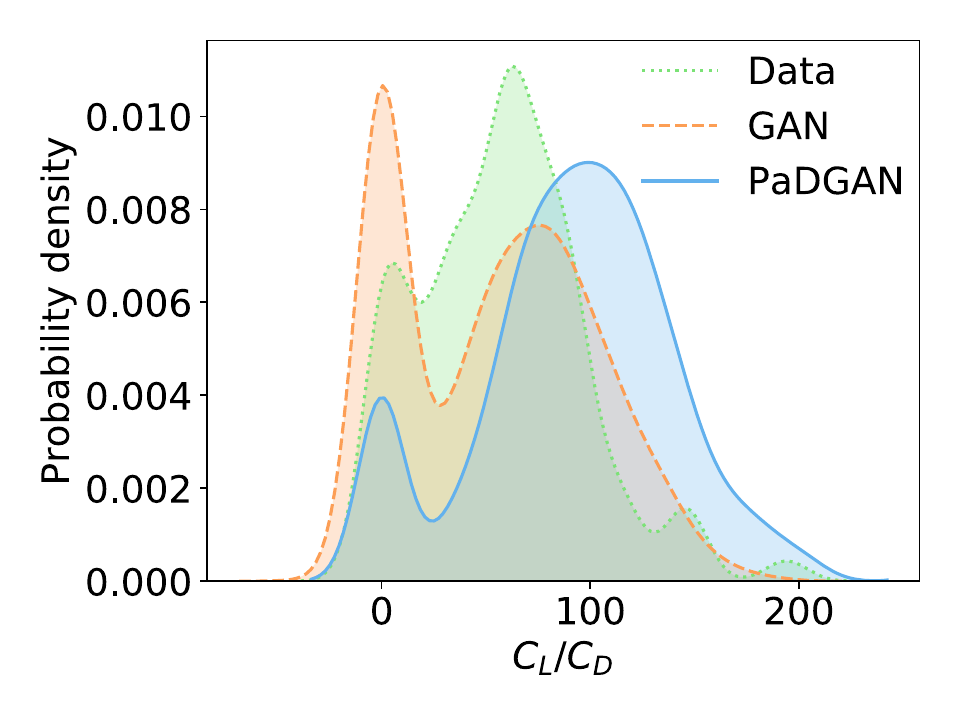}
\caption{The distribution of quality ($C_L/C_D$) for training data, vanilla GAN, and PaDGAN.}
\label{fig:airfoils_quality_dist}
\end{figure}

\begin{figure}[!ht]
\centering
\includegraphics[width=0.45\textwidth]{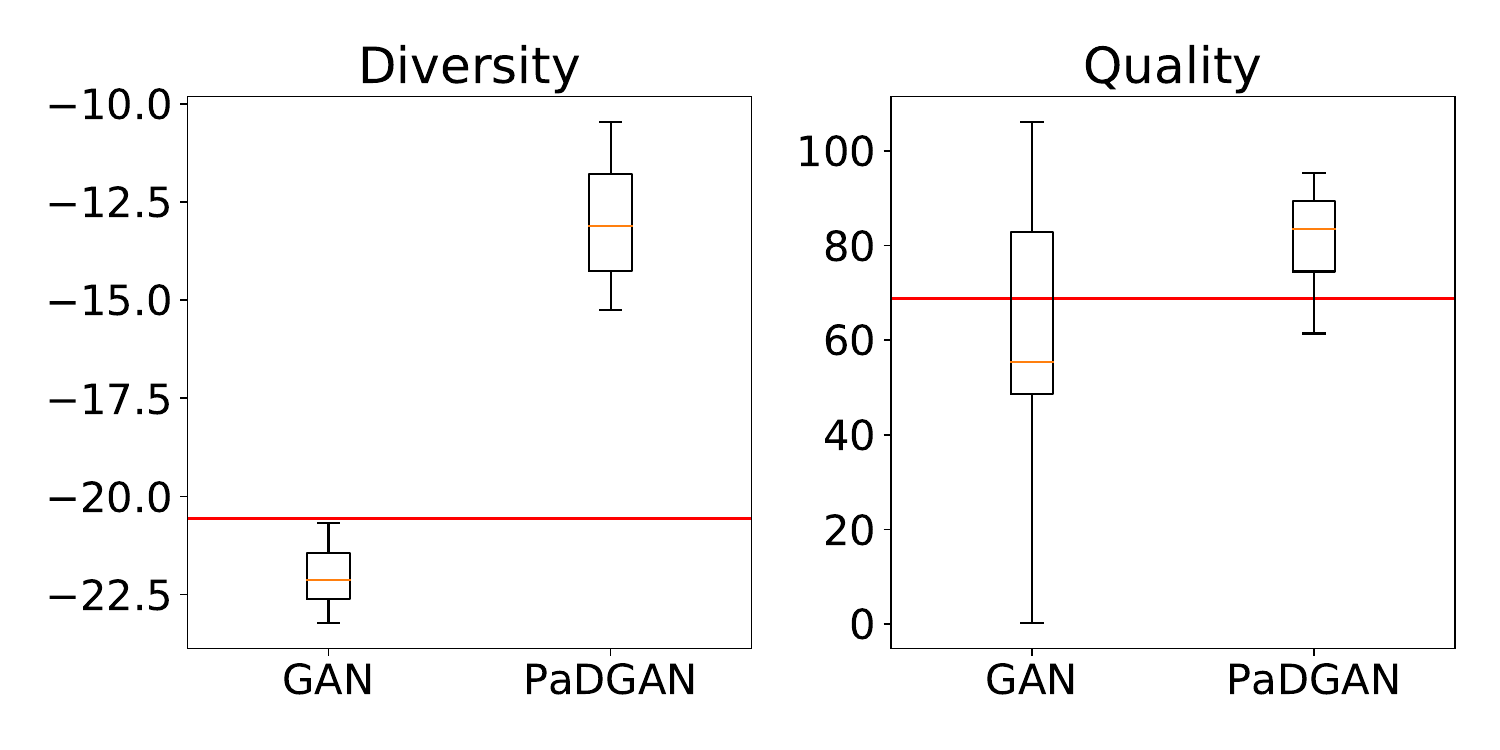}
\caption{Scores for the airfoil example. The red horizontal line denotes the diversity/quality score of the training data. The box plots show the statistics of ten models for each method.}
\label{fig:airfoil_scores}
\end{figure}

We synthesized 100 airfoil designs from a vanilla GAN and 100 from a PaDGAN, computed their quality ($C_L/C_D$ values) using XFOIL\footnote{We set $C_L/C_D=0$ when the simulation fails.}, and 
used the t-Distributed Stochastic Neighbor Embedding (t-SNE) to map these designs onto the same two-dimensional space, as shown in Fig.~\ref{fig:airfoils_tsne}. The quality is indicated by the shades of plotted designs, where dark shaded airfoils are of higher quality. We also show 100 designs from the training data in the left most figure to represent the original design space. 
Both the GAN and the PaDGAN generate realistic airfoil designs. 
We observe that the vanilla GAN (middle figure) generates a few airfoils that fill in the gaps of the training data (\ie, interpolation). However, PaDGAN discovers new high-quality designs, which are outside the boundary of the training data. We mark these regions in by ellipses in the leftmost part of Fig.~\ref{fig:airfoils_tsne}. This shows that the diversity promoting part of PaDGAN encourages it to discover new unseen design areas while the quality promoting part helps it find areas where high-quality designs are found, as is also demonstrated by synthetic example \RNum{3}.
% This promising result shows that by diversifying generated samples, PaDGAN is capable of expanding the design space towards the direction of high-quality regions. It is not surprising that the generator knows which direction to expand since it receives from the performance estimator the information of quality gradients. 
In future work, we will explore if PaDGAN can be used as a tool to assist in design discovery by generating novel high-quality designs for more complex design domains.

We show the quality (\ie, $C_L/C_D$) distributions of training data and generated designs by vanilla GAN and PaDGAN in Fig.~\ref{fig:airfoils_quality_dist}. 
We observe that the quality distribution of data has two modes (large number of samples)~\textemdash~one near 0 and one near 70. The vanilla GAN's quality distribution mimics these two modes but has a larger probability mass near 0. 
Comparing with both the training data and the vanilla GAN, PaDGAN's quality distribution has a larger mass over the higher-quality region. This shows that PaDGAN generates most samples which are of significantly higher quality than the training data.

Figure~\ref{fig:airfoil_scores} shows the statistics of quality and diversity scores over ten runs of model training. The PaDGAN's diversity score is always higher than the training data's (shown by a red horizontal line), whereas the vanilla GAN almost always has a lower diversity score than the data. The quality score of the PaDGAN has a higher mean and lower variance than the vanilla GAN. These results demonstrate the effectiveness of PaDGAN as a design exploration tool.

To show the evaluation scores in Figs.~\ref{fig:Donut2D_MixRing6_scores}-\ref{fig:ThinDonut2D_MixRing6_scores} and Fig.~\ref{fig:airfoil_scores} more clearly, we list the means and 95\% confidential intervals of all scores in Appendix~A.

%\section{Discussion and Limitations}

%Computational complexity
%  The computational complexity of eigenvalue decomposition for a matrix is $O(n^3)$. Hence the model takes longer training time for large batchsizes. 
%  This issue can be partially alleviated using faster kernel decomposition methods~\cite{chonavel2003fast} or by using a low-rank kernel approximation~\cite{gartrell2017low}.
 
%Need for a surrogate model for quality

%%%%%%%%%%%%%%%%%%%%%%%%%%%%%%%%%%%%%%%%%%%%%%%%%%%%%%%%%%%%%%%%%%%%%%
\section{Conclusion and Future Work}
 
%Estimating trade-off parameters

In this paper, we proposed a new loss function for generative models based on Determinantal Point Processes. With this loss function, we developed a new GAN model, named PaDGAN. To the best of authors' knowledge, this is the first GAN model that can simultaneously encourage the generation of diverse and high-quality designs. 
We use both synthetic and real-world examples to demonstrate the effectiveness of PaDGAN and show that by diversifying generated samples, PaDGAN expands the existing boundary of the design space towards high-quality regions.
This model is particularly useful when we want to thoroughly explore different high-quality design alternatives or discover novel solutions. For example, when performing design optimization, one may accelerate the search for global optimal solutions by sampling start points from the proposed model. Also, this method can be a tool in the early conceptual design stage to aid the creative process. It can generate new designs which are learnt from previous generations of designs, while introducing novelty and taking into account the desired quality metrics. The resultant designs can be used as inspirations to steer designers in exploring novel designs. Although we demonstrated the effectiveness of our method via a GAN-based model, the proposed framework also generalizes to other generative models like variational autoencoders and can be used for various design synthesis problems.

Note that by trying to mimic the training data, PaDGAN captures design constraints implicitly. For instance, in Fig.\ref{fig:ThinDonut2D_MixRing6_expand_density} (Example \RNum{3}), it captures the inner and outer ring of the training data and generates the majority of the points inside the two circular rings. However, we still observe a few points outside the rings, as we do not explicitly define this as a constraint boundary. To explicitly capture design constraints, one can train a differentiable classifier (\eg, a neural network-based classifier) which predicts constraint satisfaction and use it as a second discriminator. However, this approach of explicitly capturing the constraints is outside the scope of this work.

%\subsection{Broader Application outside Design}

While we developed this method for design applications, it can generalize to many other domains, where quality and coverage over a domain are needed. For example, in molecule discovery, our model can be integrated with the generative model developed by G{\'o}mez-Bombarelli~\etal{}~\cite{gomez2018automatic}, who combined a generative model with the search over latent space to generate new molecules. In 3D shape synthesis, our model can be trained on large datasets like ShapeNet and used as a recommender system within CAD software. The loss function we develop can also be integrated with human face synthesis methods, to generate new human faces, which are high quality (depending on any criteria like beauty) and from different groups (regions, race, gender, age \etc{}). Overall, the method provides a new direction of research, where generative models focus on the unbiased generation of high-quality items.

%%%%%%%%%%%%%%%%%%%%%%%%%%%%%%%%%%%%%%%%%%%%%%%%%%%%%%%%%%%%%%%%%%%%%%
% \begin{acknowledgment}
% ASME Technical Publications provided the format specifications for the Journal of Mechanical Design, though they are not easy to reproduce.  It is their commitment to ensuring quality figures in every issue of JMD that motivates this effort to have authors review the presentation of their figures.  

% Thanks go to D. E. Knuth and L. Lamport for developing the wonderful word processing software packages \TeX\ and \LaTeX. We would like to thank Ken Sprott, Kirk van Katwyk, and Matt Campbell for fixing bugs in the ASME style file \verb+asme2ej.cls+, and Geoff Shiflett for creating 
% ASME bibliography stype file \verb+asmems4.bst+.
% \end{acknowledgment}

%%%%%%%%%%%%%%%%%%%%%%%%%%%%%%%%%%%%%%%%%%%%%%%%%%%%%%%%%%%%%%%%%%%%%%
% The bibliography is stored in an external database file
% in the BibTeX format (file_name.bib).  The bibliography is
% created by the following command and it will appear in this
% position in the document. You may, of course, create your
% own bibliography by using thebibliography environment as in
%
% \begin{thebibliography}{12}
% ...
% \bibitem{itemreference} D. E. Knudsen.
% {\em 1966 World Bnus Almanac.}
% {Permafrost Press, Novosibirsk.}
% ...
% \end{thebibliography}

% Here's where you specify the bibliography style file.
% The full file name for the bibliography style file 
% used for an ASME paper is asmems4.bst.
\bibliographystyle{unsrt}

% Here's where you specify the bibliography database file.
% The full file name of the bibliography database for this
% article is asme2e.bib. The name for your database is up
% to you.
\bibliography{arxiv}

%%%%%%%%%%%%%%%%%%%%%%%%%%%%%%%%%%%%%%%%%%%%%%%%%%%%%%%%%%%%%%%%%%%%%%
\newpage
\appendix       %%% starting appendix
\section*{Appendix A: Table of Evaluation Metrics}

We list the means and 95\% confidence intervals of all evaluation scores (Figs.~\ref{fig:Donut2D_MixRing6_scores}-\ref{fig:ThinDonut2D_MixRing6_scores} and Fig.~\ref{fig:airfoil_scores}) in Table~\ref{tab:scores}. It shows that PaDGAN received best overall score for all cases, and atleast the second best score for both diversity and quality in all four examples.

\begin{table*}[hbt!]
\centering
\caption{Diversity and Quality Scores for all experiments. The best score for each example is marked by $^{*~}$ symbol and the second best scores by $^{**}$ symbol.}
\label{tab:scores}       % Give a unique label
% For LaTeX tables use
\begin{tabular}{llrrr}
% \begin{tabular}{p{0.2cm}p{1cm}
% >{\raggedleft\arraybackslash}p{3.27cm}
% >{\raggedright\arraybackslash}p{0.05cm}
% p{3.5cm}p{3.5cm}}
& & & &\\ % put some space after the caption
\hline\noalign{\smallskip}
& Model & Diversity Score & Quality Score & Overall Score  \\
\noalign{\smallskip}\hline\noalign{\smallskip}
\multirow{4}{*}{\begin{sideways} Example~\RNum{1} \end{sideways}} 
& GAN & $-69.0885 \pm 27.4357^{~~}$  & $0.5826 \pm 0.0620^{~~}$  & $1.0678 \pm 0.2938^{**}$ \\[5pt]
& GAN$_D$ & $-53.6183 \pm 0.3601^{*~}$& $0.4672 \pm 0.0162^{~~}$  & $0.9060 \pm 0.0414^{~~}$ \\[5pt]
& GAN$_Q$ & $-152.1945 \pm 77.5582^{~~}$  &$0.9131 \pm 0.1065^{*~}$  & $0.5913 \pm 0.5875^{~~}$ \\[5pt]
& PaDGAN & $-57.2489 \pm 1.16202^{**}$  & $0.6955 \pm 0.0269^{**}$  & $1.4897 \pm 0.0624^{*~}$ \\[5pt]
\hline\noalign{\smallskip}
\multirow{4}{*}{\begin{sideways} Example~\RNum{2} \end{sideways}} 
& GAN  & $-68.2451 \pm 3.9807^{~~}$  & $0.5004 \pm 0.0447^{~~}$  & $0.9752 \pm 0.0531^{**}$ \\[5pt]
& GAN$_D$ & $-61.8685 \pm 0.7375^{*~}$ & $0.4811 \pm 0.0239^{~~}$  & $0.9473 \pm 0.0269^{~~}$ \\[5pt]
& GAN$_Q$ & $-197.0243 \pm 119.0978^{~~}$  & $0.9914 \pm 0.0280^{*~}$  & $0.4261 \pm 0.8780^{~~}$ \\[5pt]
& PaDGAN & $-65.5501 \pm 1.53222^{**}$  & $0.7176 \pm 0.0344^{**}$  & $1.2117 \pm 0.0320^{*~}$ \\[5pt]
\hline\noalign{\smallskip}
\multirow{4}{*}{\begin{sideways} Example~\RNum{3} \end{sideways}} 
& GAN & $-93.1237 \pm 67.4991^{~~}$ & $0.6219 \pm 0.0847^{~~}$  & $0.9019 \pm 0.8333^{~~}$ \\[5pt]
& GAN$_D$ & $-63.5478 \pm 0.4637^{*~}$ & $0.5560 \pm 0.0152^{~~}$  & $1.0811 \pm 0.0468^{**}$ \\[5pt]
& GAN$_Q$ & $-150.3896 \pm 65.4173^{~~}$  & $0.8492 \pm 0.1219^{*~}$  & $0.6356 \pm 0.6366^{~~}$ \\[5pt]
& PaDGAN & $-63.9063 \pm 1.76832^{**}$  & $0.6968 \pm 0.0248^{**}$  & $1.4993 \pm 0.0624^{*~}$ \\[5pt]
\hline\noalign{\smallskip}
\multirow{2}{*}{\begin{sideways} Airfoil \end{sideways}} 
& GAN & $-22.0326 \pm 1.4982^{**}$  & $56.1703 \pm 66.4480^{**}$ & N/A \\[5pt]
& PaDGAN & $-12.5455 \pm 4.8970^{*~}$& $76.0670 \pm 43.9802^{*~}$ & N/A \\[5pt]
\noalign{\smallskip}\hline
\end{tabular}
\end{table*}

\end{document}